\documentclass{article}

\usepackage{microtype}
\usepackage{graphicx}
\graphicspath{{figures/}{../figures/}}
\usepackage{subcaption}
\usepackage{booktabs}
\usepackage{tikz}

\usepackage{hyperref}

\usepackage[accepted]{icml2026}

\usepackage{amsmath}
\usepackage{amssymb}
\usepackage{mathtools}
\usepackage{amsthm}

\usepackage[capitalize,noabbrev]{cleveref}

\makeatletter
\newcommand{\floatlabel}[1]{\cref@label{#1}}
\makeatother

\theoremstyle{plain}

\theoremstyle{definition}

\theoremstyle{remark}

\renewcommand{\texttt}[1]{#1}
\icmltitlerunning{When Offline Selectors Cannot Beat the Best Single Model}

\begin{document}

\twocolumn[
\icmltitle{When Offline Selectors Cannot Beat the Best Single Model:\\
  A Diagnostic Study on edX Dropout Prediction}

\begin{icmlauthorlist}
\icmlauthor{Tyler Crosse}{gt}
\icmlauthor{Alan Nadelsticher Ruvalcaba}{gt}
\icmlauthor{Dustin Khang LeDuc}{gt}
\icmlauthor{Thomas Trask}{gt}
\icmlauthor{Nicholas Lytle}{gt}
\icmlauthor{David Joyner}{gt}
\end{icmlauthorlist}

\icmlaffiliation{gt}{Georgia Institute of Technology}

\icmlcorrespondingauthor{Tyler Crosse}{tylerscottcrosse@gmail.com}

\icmlkeywords{offline reinforcement learning, meta-learning, model selection,
  distribution shift, diagnostic analysis, educational data mining}

\vskip 0.3in
]

\printAffiliationsAndNotice{
  Accepted to the ICML 2026 Workshop on Decision-Making from Offline Datasets to Online Adaptation (DEMO).
  \\
}

\begin{abstract}
Different predictors often excel on different inputs, so picking the
best one per instance promises higher accuracy than committing to a
single model. In practice, selectors trained from logged data
routinely fail to beat the strongest single predictor. Three causes
typically go unseparated before more tuning is applied: a mismatched
learner, a state that does not predict which model wins, or
buffer-to-deployment label shift.

A three-stage diagnostic rules them out on a shared buffer. Stage~1
estimates a local ceiling on oracle recovery from $k$-NN label
consistency. Stage~2 asks whether paired BC and offline-RL learners
(BC, DQN, and CQL across penalty weights) reach that ceiling.
Stage~3 ablates the selector state to test whether richer features
would raise it. The combined verdict points to the most promising
next step: tuning the learner, redesigning the state, or collecting
new data.

We apply it to selecting among five dropout-prediction models on edX
clickstream data. Across 16 windows, the oracle beats the strongest
single base model by 9.7 accuracy points on average, yet BC, DQN, and
CQL land in the same test-accuracy band below it (robust to a tenfold
buffer sweep and $N{=}2{,}000$ held-out examples). The bottleneck is
local representational ambiguity: CQL closes the imitation gap
without a deployment gain (not conservatism), regret clusters tightly
across learners (not tie-breaking), and the three learners converge
on test accuracy (not shift). The next iteration should change the
state or collect new data, not tune the offline learner further.
\end{abstract}

\section{Introduction}
\label{sec:intro}

Decision-making from offline datasets is central to a growing class of
applications where online interaction is expensive, slow, or ethically
constrained, including scientific discovery, engineering design,
healthcare, and education~\cite{levine:offline}. The data take many
forms (logged demonstrations, past trajectories, recorded
interactions). A recurring special case is meta-learning over a pool of
base models, where a policy chooses the best base model for each
instance instead of committing to a single
predictor~\cite{rice:selection,cruz:des,cruz:metades}. The setup is
attractive because individual base models can exhibit context-dependent
competence, and prior educational prediction studies commonly compare
several plausible model classes on clickstream and
virtual-learning-environment
data~\cite{liu:clickstream,taylor:stopout,casado:metalearning}. In
practice, offline meta-learning routinely underperforms strong static
baselines, and the reasons are opaque. The weakness could be
\emph{algorithmic} (offline
RL's conservatism~\cite{kumar:cql} or reward misspecification),
\emph{representational} (the state does not contain the information
needed to predict which model wins), or \emph{distributional} (the
oracle-label distribution available offline differs from the one
induced at evaluation). Existing evaluations rarely separate these
causes. As a result, practitioners tune algorithms that cannot fix
representation failures, engineer features that add zero marginal
signal, or attribute failure to shift without quantifying it.

We rule out the three hypotheses in turn with three diagnostics on a
shared buffer (\Cref{fig:diagnostic-protocol}).
\textbf{Stage~1} measures local label consistency. The
$k$-nearest-neighbor consistency quantifies how often nearby states in
the buffer share the same oracle action, and a held-out $10$-NN
selector calibrates how that local ambiguity translates into test-time
imitation. \textbf{Stage~2} separates algorithmic failure from
representational failure. We train a supervised behavioral-cloning
policy with hard-label cross-entropy and an offline Deep
Q-Network~\cite{mnih:dqn} on the same buffer. If both fail by similar
margins, the bottleneck is more plausibly shared, pointing to
representation or distribution rather than algorithm choice.
\textbf{Stage~3}
isolates the marginal value of features. State ablations test whether
the full behavioral state improves over the base-model probability
vector, and whether disagreement-derived transforms of that vector add
anything further.

The three stages constrain each other. Stage~1's local-consistency
ceiling is what makes Stage~2's learner-agreement gap interpretable:
learners converging close to a low ceiling implicates the
representation, while learners separating below a high ceiling
implicates the algorithm. Stage~3 then tests whether reachable
features could raise the ceiling. Run together on a shared buffer, the
three checks identify which intervention---more tuning, richer
features, or upstream data collection---matters next. Each individual
diagnostic ($k$-NN consistency, paired BC/RL ablation, feature
ablation, and total-variation buffer-to-test shift) has antecedents in
prior work~\cite{cruz:des,ko:des,kumar:cql}; the contribution is the
joint reading they enable.

\begin{figure}[ht]
\centering
\includegraphics[width=\columnwidth]{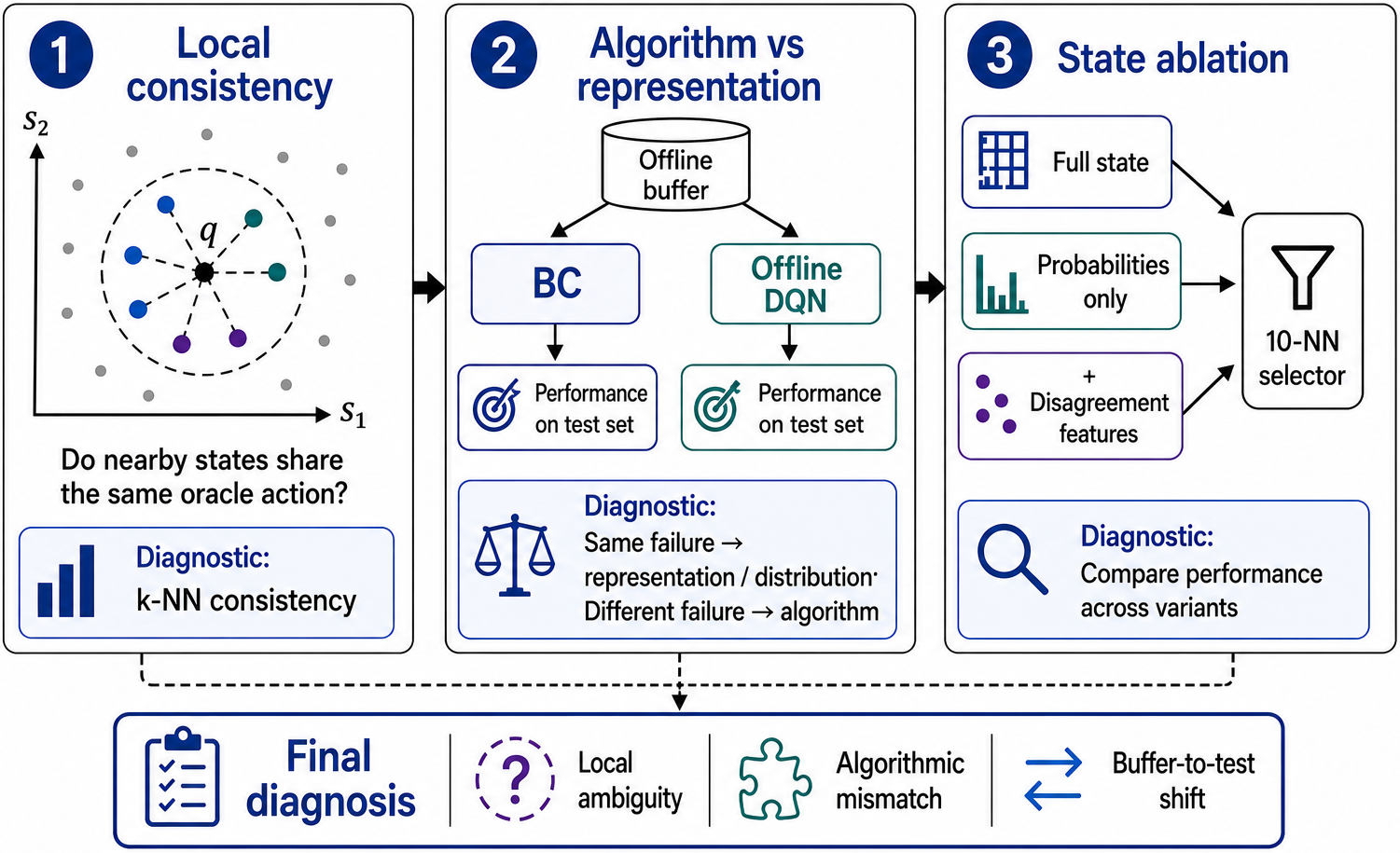}
\caption{Three-stage diagnostic protocol for offline model selection. The
same offline buffer is first used to measure local oracle consistency, then
to compare algorithm-specific and shared failure modes across BC, offline
DQN, and CQL at three penalty weights, and finally to ablate the selector
state to test whether additional feature groups provide marginal value
beyond base-model probabilities.}
\floatlabel{fig:diagnostic-protocol}
\end{figure}

We apply the protocol to a concrete offline decision-making task:
selecting among five static dropout-prediction models for MOOC and
in-person computer science students on edX clickstream data. The task
combines abundant observational data (84.5M events across $223{,}505$
student-course pairs) with ethically constrained online
experimentation, because any selected model eventually triggers a human
intervention. Across 16 observation/prediction-window configurations, the
per-instance oracle beats the strongest single base model by 9.7
accuracy points on average (range 4.5--15.5), yet no learned selector
recovers that headroom on held-out accuracy. On the main
$(14\text{d},14\text{d})$ configuration, every learner sits within
$\pm 0.01$ of $0.748$ test accuracy, below the $0.762$ static
reference, while the local-consistency diagnostic is only
$0.388 \pm 0.010$ (\Cref{tab:main-results}). State ablations show that
probabilities-only BC nearly matches the full state, and that
disagreement-derived transforms do not materially improve it.
Buffer-to-test shift is small in aggregate (marginal
$d_{\mathrm{TV}} = 0.063 \pm 0.011$) but locally substantial
($\mathbb{E}_s[d_{\mathrm{TV}}] \approx 0.29$ on the main
configuration). The procedure converts an opaque negative result into a
verdict on whether the next iteration should target the learner, the
features, or the data-collection pipeline. Our contributions are:

\begin{itemize}
  \item \textbf{C1.} A combined diagnostic procedure assembled from
  established checks ($k$-NN consistency, paired BC/RL ablation, state
  ablation, and marginal/conditional $d_{\mathrm{TV}}$) for deciding
  whether offline data is sufficient before further offline tuning or
  online adaptation. We apply it as a single-task case study; whether
  it generalizes to other settings is left to future work.
  \item \textbf{C2.} An empirical case study on selecting among five
  pre-trained dropout-prediction models on edX clickstream data. Across
  BC, offline DQN, and CQL at three penalty weights, no learned
  selector beats the strongest single base model on held-out accuracy
  despite a 9.7-point per-instance oracle gap. Mechanism checks weigh
  against three candidate causes: algorithmic conservatism (CQL closes
  the imitation gap without a deployment gain), hard-label tie-breaking
  (regret clusters in $[0.089, 0.101]$ across BC/DQN/CQL while oracle
  agreement spreads over $0.36$ to $0.52$), and buffer-to-test marginal
  shift. The remaining candidate, consistent with the diagnostic
  readout, is local label ambiguity. Richer offline
  encodings of the same probability vector (disagreement-derived
  transforms and the full 38-d state) do not measurably improve
  deployment accuracy over the 5-d probability subspace alone, and the
  result is insensitive to training-buffer size.
\end{itemize}

Shared-failure patterns plausibly arise in other offline meta-learning
settings that combine pre-trained predictors, including drug-response
prediction, content recommendation, and offline hyperparameter
selection. Whether the diagnostic procedure transfers to those
settings is an open question we hope to test.

\section{Related Work}
\label{sec:related}

This paper sits at the intersection of offline decision-making, dynamic
model selection, and educational outcome prediction. Offline
RL~\cite{levine:offline} must cope with support mismatch between the
data distribution and the policy desired at deployment, while
contextual-bandit and offline-policy-evaluation work in education provide
closely related decision framings~\cite{lan:bandits,mandel:ope}.
Conservative or behavior-constrained methods such as CQL, IQL, BCQ, BRAC,
and BEAR~\cite{kumar:cql,kostrikov:iql,fujimoto:bcq,wu:brac,kumar:bear}
aim to reduce
offline-RL failures caused by distribution shift, extrapolation error,
or out-of-distribution action evaluation, but they do not by themselves
resolve ambiguity in which action is optimal from the available state.
Our contribution is therefore diagnostic. We quantify when the offline
state and label construction are too ambiguous for the selectors we
evaluate---a paired BC/offline-DQN/CQL family on a shared buffer---to
recover oracle headroom, and we identify this as a failure mode that
conservatism alone does not resolve.

The per-instance selection problem traces back to algorithm
selection~\cite{rice:selection} and dynamic classifier
selection~\cite{cruz:des,ko:des,cruz:metades}, where different models
dominate in different regions of feature space. In education, related
work spans clickstream-based MOOC dropout
prediction~\cite{dalipi:review,salleh:randomforest,taylor:stopout,nagrecha:temporal},
e-learning dropout prediction with model
combinations~\cite{lykourentzou:ensemble}, higher-education
dropout prediction with administrative and LMS data~\cite{berens:early},
and meta-learning for student-performance
prediction~\cite{casado:metalearning}. We build on that observation but
focus on diagnosing why offline adaptive selection fails even when
oracle headroom exists. Each diagnostic in our procedure has
antecedents elsewhere. The $k$-NN label consistency is closely related
to neighborhood-purity measures used in dynamic classifier
selection~\cite{cruz:des,ko:des}, paired BC-versus-offline-RL
ablations are a routine offline-RL diagnostic pattern, and feature ablations are
standard ML practice. The contribution is the combined empirical application on this task,
including the conservatism check against CQL~\cite{kumar:cql} and the
conditional buffer-to-test shift estimate. The setup also resembles
algorithm selection from ex post oracle-labeled buffers more than a
full logged-bandit benchmark with historical action
propensities~\cite{dudik:doublyrobust,swaminathan:crm}.\footnote{We
therefore treat offline RL here as one comparator family among
several, not as the sole framing.}

The paper also contributes to evaluation protocol. Offline-learning
benchmarks often report final policy quality without first quantifying
whether the dataset contains actionable oracle headroom, whether simple
local predictors can recover the relevant action labels, or whether
distribution shift is large enough to dominate the result. This protocol
turns those hidden assumptions into explicit measurements.

\section{Problem Formulation}
\label{sec:formulation}

We formalize dynamic model selection as a contextual bandit. The
oracle-label distribution is induced explicitly by the buffer
construction. This formulation is sufficient for the one-step decision
problem studied here and lets us separate algorithmic failure from
distributional failure later in the protocol.

\paragraph{Decision task.}
We are given a pool of pre-trained base classifiers
$\mathcal{M} = \{m_1, \dots, m_K\}$. Here $K{=}5$ (logistic regression,
random forest, gradient boosting, calibrated random forest, and a
stacking ensemble). For each student-course pair, a 14-day observation
window is summarized as a state $s \in \mathcal{S} \subseteq
\mathbb{R}^d$, and the task is to select a single model $a \in
\mathcal{A} = \{1, \dots, K\}$ whose prediction for that sample is used
at deployment.

\paragraph{State and reward.}
The state vector concatenates 28 engineered behavioral features (volume,
consistency, temporal patterns, trends; see \Cref{sec:setup}) with a
binary modality flag, the $K$-dimensional vector of base-model dropout
probabilities on that sample, their mean, and a 3-way one-hot bin
derived from that mean, yielding the 38-dimensional state used in the
main experiments. The deployment metric is the selected model's
zero-one correctness on the 14-day prediction label,
$R_{\mathrm{eval}}(s, a) = \mathbb{I}[m_a(s) \text{ predicts the true label}]$.
The offline DQN in the body is trained with the canonical oracle-match
reward $R(s,a) = R_{\mathrm{eval}}(s,a)$ on the buffer. The
log-probability shaping variant used in earlier drafts is preserved as a
sensitivity row in Appendix~\ref{app:dqn-marginals}. Throughout,
\emph{oracle agreement} is a label-imitation diagnostic against a
single argmax action, \emph{test accuracy} is the deployment-facing
metric, and \emph{regret} (\Cref{subsec:res-regret}) is the fraction of
test samples on which the policy selects an incorrect model when at
least one correct model exists.

\paragraph{Contextual-bandit reduction.}
Because each sample's observation window is fixed and does not evolve
under the agent's choice, within-sample state transitions are
degenerate. We set the discount factor $\gamma{=}0$, reducing the MDP
to a contextual bandit~\cite{lan:bandits}. The Deep $Q$-Network used
in \Cref{sec:cascade} retains its full architecture but learns only
the immediate action-value $Q(s, a)$, with no bootstrapping term.

\paragraph{Offline buffer and buffer oracle distribution.}
The buffer $\mathcal{B} = \{(s_i, a^\star_i, r_i)\}_{i=1}^N$ is built by
4-fold cross-validation on the training split. For each fold, base
models are fit on the other three folds and scored on the held-out
fold. The \emph{oracle action} is the hard label
\begin{equation}
  a^\star_i \;=\; \arg\max_{a \in \mathcal{A}}
  \left[y_i\,p_a(s_i) + (1-y_i)\,(1-p_a(s_i))\right],
\end{equation}
that is, the base model assigning the highest probability to the true
class. The induced distribution $\pi_\beta(a \mid s)$ is the
categorical distribution over these oracle labels. It is reconstructed
from cross-validation, so it has no observed action propensities and
should not be read as a historical logging policy. Every offline
method in this paper (held-out $10$-NN selection, behavioral cloning,
offline DQN, and the state-ablation variants) sees $\mathcal{B}$ and
nothing else.

\paragraph{Evaluation oracle and buffer-to-test shift.}
At test time we compute the \emph{evaluation oracle}
$\pi^\star(a \mid s)$ by the same argmax-over-base-models construction
on the held-out 200-sample test set. The quantity the offline agent
must close is the total-variation distance between the marginal action
distributions:
\begin{equation}
  \begin{aligned}
    d_{\mathrm{TV}}\!\left(\pi_\beta, \pi^\star\right)
    &= \tfrac{1}{2} \sum_{a \in \mathcal{A}} \\
    &\qquad \bigl| \mathbb{E}_{s \sim \mathcal{B}}\,\pi_\beta(a \mid s)
      - \mathbb{E}_{s \sim \mathcal{D}_{\mathrm{test}}}\,
      \pi^\star(a \mid s) \bigr|.
  \end{aligned}
  \label{eq:dtv}
\end{equation}
\Cref{sec:results} reports this quantity per window configuration and
shows it is substantial. The CV-induced buffer concentrates probability
mass on actions that are not dominant at test time, so an offline agent
whose action distribution tracks $\pi_\beta$ will, by construction,
place mass on actions that $\pi^\star$ does not, with the gap
lower-bounded by $d_{\mathrm{TV}}(\pi_\beta, \pi^\star)$. We treat
$d_{\mathrm{TV}}(\pi_\beta,\pi^\star)$ as a buffer-to-test oracle-shift
diagnostic, since it is not a full logged-policy support estimate.

\paragraph{Off-policy evaluation framing.}
The setup doubles as an off-policy evaluation problem, with $\pi_\beta$
the behavior policy and $\pi^\star$ the target. Our test-accuracy and
regret reports (\Cref{subsec:res-regret}) provide a direct
deployment-value estimate for the selectors we evaluate, in lieu of
importance-weighted OPE. Importance
weighting~\cite{dudik:doublyrobust,swaminathan:crm} would require
action propensities, which the CV reconstruction does not provide.
\Cref{sec:discussion} returns to why conservative offline-RL methods
do not close the failure mode our diagnosis uncovers.

\section{Three-Stage Diagnostic Protocol}
\label{sec:cascade}

The protocol answers three questions in sequence. (1) Are the oracle's
selections locally consistent in the current state representation? (2)
If so, do a supervised and a reinforcement-learning approach succeed or
fail together (pointing to representation or distribution) or
differently (pointing to algorithm)? (3) Which state components carry
signal beyond what the base-model probabilities already encode? Each
stage produces a compact diagnostic. A reader who runs Stage~1 alone
can already tell whether the state is locally ambiguous before
investing in heavier learners.

\subsection{Stage 1: Local label consistency via $k$-nearest neighbors}
\label{subsec:cascade-stage1}

If neighboring states in the buffer disagree on the oracle action, no
local selector can rely on a stable neighborhood rule. We measure this
ambiguity directly. For each buffer sample
$(s_i, a^\star_i) \in \mathcal{B}$, we locate its $k$ nearest neighbors
in state space under the Euclidean metric on the standardized
$d$-dimensional state and compute the \emph{consistency}
\begin{equation}
  c_i \;=\; \tfrac{1}{k} \sum_{j \in \mathcal{N}_k(s_i)} \mathbb{I}\!\left[
    a^\star_j = a^\star_i \right],
  \label{eq:knn}
\end{equation}
then average $\bar c = \mathbb{E}_i[c_i]$ over the buffer. We use
$k{=}10$ throughout. Smaller $k$ introduces sampling noise; larger $k$
washes out local structure. The quantity $\bar c$ is a diagnostic,
since it is not a formal upper bound on downstream BC or DQN
performance. Low values indicate that nearby states often map to
different oracle actions, so any local state-conditioned selector must
resolve substantial ambiguity. To calibrate this diagnostic against an
actual predictor, we also evaluate a held-out $10$-NN selector that
predicts the oracle action of each test sample from its nearest
training-buffer neighbors in the same standardized state space. We
report its oracle agreement and test accuracy alongside $\bar c$.

\subsection{Stage 2: Algorithm versus representation via paired ablation}
\label{subsec:cascade-stage2}

A low local-consistency diagnostic does not by itself distinguish
algorithmic failure from representational failure. To separate them, we
train two policies on the same buffer using architectures that would
fail for different reasons if the bottleneck were algorithmic. The
first is a supervised behavioral cloner that does not require reward
engineering. The second is an offline $Q$-learner that does. If both
fail by similar margins, the bottleneck is more plausibly shared,
pointing to representation or distribution.

The behavioral-cloning policy $\pi^{\mathrm{BC}}(a \mid s)$ is a
two-layer MLP ($d \to 64 \to K$) trained with cross-entropy loss
against the hard oracle labels from the 4-fold cross-validation buffer,
for 30 epochs with dropout $0.2$ and weight decay $10^{-4}$. This is a
vanilla supervised multi-class classifier, so any failure here cannot
be attributed to reward sparsity, bootstrapping instability, or
target-network dynamics.

The Deep $Q$-Network~\cite{mnih:dqn} is a larger MLP
($d \to 128 \to 64 \to K$) trained with the one-step bandit objective
($\gamma{=}0$), Adam, soft target updates with rate $\tau{=}0.05$, and
a replay buffer of the full offline $\mathcal{B}$. We use the canonical
$\{0,1\}$ oracle-match reward
$R(s,a) = \mathbb{I}[m_a(s) \text{ predicts the true label}]$, which
matches the deployment metric and avoids the policy collapse that
arises under log-probability shaping (see
Appendix~\ref{app:dqn-marginals} for the reward-sensitivity analysis).
The network is trained for 50 epochs on the same buffer used by BC.

To control for offline-RL conservatism, we also train a Conservative
$Q$-Learning (CQL)~\cite{kumar:cql} variant on the same buffer,
network, and reward. CQL adds the standard penalty
$\alpha\,\mathbb{E}_s\!\left[\log\sum_{a'} \exp Q(s,a') -
Q(s, a^\star)\right]$ to the Bellman loss, suppressing $Q$-values for
actions outside the buffer's support. We sweep
$\alpha \in \{0.1, 1.0, 5.0\}$. We anchor the second term on the hard
buffer-oracle action $a^\star$ instead of
$\mathbb{E}_{a \sim \hat\pi_\beta(a \mid s)}[Q(s,a)]$; under the
deterministic CV-induced label distribution this collapses to the
standard form and avoids estimating a stochastic behavior policy.
Because $\gamma{=}0$, the conservatism term is the only thing
distinguishing CQL from DQN here, which isolates the contribution of
pessimism from any bootstrapping effect.

For each policy we report \emph{oracle agreement} (the fraction of
test samples on which the policy's selected action matches the hard
evaluation-oracle label) and \emph{test accuracy} (the accuracy of the
selected base model's prediction on those samples). The first
measures how well the policy imitates the oracle; the second measures
whether imitation translates into useful predictions. A low oracle
agreement combined with near-reference accuracy is consistent with a
policy defaulting to a strong single model regardless of $s$.

\subsection{Stage 3: Feature marginal value via state ablations}
\label{subsec:cascade-stage3}

If the 28 engineered behavioral features carry unique signal for model
selection (information not already encoded in the 5 base-model
probabilities) \emph{that the meta-learner can exploit}, we would
expect removing them to reduce held-out accuracy. We construct a second pair of policies (BC and a
probabilities-only MLP baseline) whose input is the 5-dimensional
probability vector alone, training them on the same buffer with matched
optimizer settings. The difference in test accuracy between full-state
($d{=}38$) and probabilities-only ($d{=}5$) variants isolates the
marginal contribution of the engineered features to the meta-learning
task, separately from their well-established contribution to the
\emph{base-model} prediction task. (The base models still use the full
28 behavioral features during their own training.) A null difference
is consistent with the 28 features being redundant for model selection
under our learners and sample size, despite their established value
for the base-model prediction task; it does not, on its own, rule out
an information-bearing signal that a different architecture or larger
buffer could exploit.

We then augment the 38-dimensional state with 13 disagreement-derived
transforms of the same base-model probability vector. These include
the probability standard deviation, all pairwise absolute differences,
the predictive entropy of the mean probability, and the top-1/top-2
margin. They serve as a representation-bias test. They are also
deterministic transforms of quantities already in the state and so do
not add independent information; a failure to improve held-out
accuracy is evidence against this particular tweak, but not against
any future augmentation.

\section{Experimental Setup}
\label{sec:setup}

This section specifies the dataset, sampling, features, base models,
and buffer construction underlying every experiment in
\Cref{sec:results}. Each choice is reproducible from the released
configuration. The only source of variance across runs is the random
seed controlling sample draws, model initialization, and the stochastic
cross-validation partition.

\subsection{Dataset and inclusion criteria}

We use edX clickstream data from 52 offerings of the same introductory
computer science course, covering $223{,}505$ student-course pairs and
84.5~million events. We retain learners with at least 10 events, at
least 3 active days, and at least 28 days of activity span, yielding
$34{,}303$ qualifying pairs from $21{,}990$ students. For each
experimental configuration we draw a stratified sample of $1{,}000$
student-course pairs, so that all 16 window configurations and 5-seed
sweeps are evaluated under the same per-configuration budget. The
resulting diagnostics are estimates under this sampled regime, not
full-corpus limits. The corpus mixes 38 MOOC offerings and 9 in-person
sections, so the state includes a binary modality flag. Additional
cohort and feature details are in Appendix~\ref{app:feature-setup} and
\Cref{tab:app-feature-categories}.

\subsection{Windows and labels}

For each sampled pair we construct an observation window followed by a
prediction window. The main configuration is
$(14\text{d},14\text{d})$. Dropout is defined as no events in the
prediction window, giving a 33.7\% positive rate. We also evaluate the
full $4\times4$ grid with
$T_{\mathrm{obs}}, T_{\mathrm{pred}} \in \{7, 14, 21, 28\}$ days. We
center on $(14\text{d},14\text{d})$ because shorter horizons are
noisier and longer horizons are less actionable for intervention.

\subsection{Engineered features}
\label{subsec:features}

From each observation window we extract 28 aggregate behavioral features
covering activity volume, content ratios, consistency, temporal
placement, trends, and concentration of activity. All features are
standardized for the linear components of the pipeline. The feature
taxonomy and examples are in \Cref{tab:app-feature-categories}.

\subsection{Base models}
\label{subsec:base-models}

We train five base classifiers on the full training set using all 28
engineered features plus the binary modality flag.
\textbf{LR} is L2-regularized logistic regression.
\textbf{RF} is a random forest (400 trees, depth 16, min-split 2).
\textbf{GB} is gradient boosting (300 trees, lr 0.05, depth 3).
\textbf{Calibrated RF} is an isotonic-CV calibrated random forest.
\textbf{Stacking}~\cite{wolpert:stacking} is a logistic-regression
meta-learner over LR, RF, and GB. Each model outputs a scalar dropout probability
$p_i(s) \in [0, 1]$. The meta-learning action set
$\mathcal{A} = \{1, \ldots, 5\}$ corresponds to these five models.

\subsection{Buffer construction and evaluation protocol}

We split each $1{,}000$-pair sample into stratified train ($800$) and
test ($200$) partitions. The offline buffer is built from 4-fold
cross-validation on the training split, yielding out-of-fold base-model
probabilities and hard oracle actions without leakage. The main state
has 38 dimensions: 28 behavioral features, 1 modality flag, 5
base-model probabilities, 1 mean probability, and a 3-way one-hot risk
bin. The disagreement augmentation adds 13 deterministic transforms of
the probability vector, yielding a 51-dimensional augmented state. When
we report the \emph{best static} or \emph{strongest single-model}
reference, it is the single base model with the highest accuracy on
that held-out test split. We report it as a hindsight reference, not as
a deployable model-selection rule.

All reported numbers are mean $\pm$ standard deviation across 5 random
seeds, controlling the sample draw, train/test split, and CV partition.
We report test accuracy and oracle agreement throughout;
Appendix~\ref{app:training-details} provides training details and the
bootstrap protocol used for paired intervals. The main tables combine
three uncertainty sources at once: resampling of examples, resplitting
of train and test, and stochastic model training.

A sample-size sensitivity analysis on this main configuration, holding
the test partition fixed and subsampling only the training buffer, is
reported in \Cref{subsec:res-sample-size}. Code, experiment
configurations, and aggregated CSV outputs underlying every table and
figure are released at
\url{https://anonymous.4open.science/r/gtedm-icml26}.\footnote{Anonymized
for double-blind review. The link will be replaced with the public
repository at camera-ready.}

\section{Results}
\label{sec:results}

We first establish the amount of oracle headroom available, then
summarize the three-stage diagnostic on the main configuration, and
finally quantify how much buffer-to-test oracle shift remains after
that diagnosis. \Cref{fig:main-acc-bunching} gives a compact visual
summary of the main case; Appendix~\ref{app:main-case} reports the raw
sample counts and action marginals behind it.

\subsection{Oracle headroom exists across window configurations}
\label{subsec:res-smoothness}

\begin{figure}[t]
\centering
\includegraphics[width=\columnwidth]{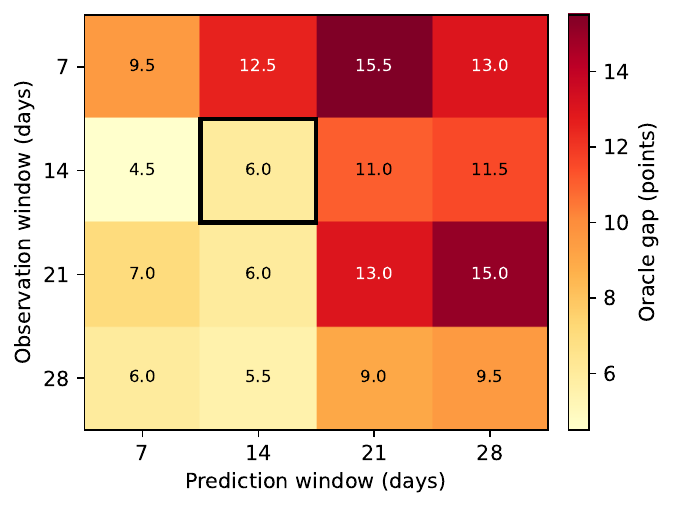}
\caption{Oracle gap across the 16 observation/prediction window
configurations. Values are absolute accuracy gains of the per-instance
oracle over the strongest single-model reference on the same test split. The
highlighted $(14\text{d},14\text{d})$ configuration lies near the middle of
the observed oracle-gap range while preserving a plausible intervention
horizon. Full window-level selector results are reported in
\Cref{tab:app-full-window-results}.}
\floatlabel{fig:oracle-gap}
\end{figure}

Across the 16 window configurations, the oracle improves on the
strongest single base model by 4.5 to 15.5 points, with a mean gap of
9.7 points. On the main $(14\text{d},14\text{d})$ configuration, the
strongest single base model reaches $0.762 \pm 0.024$ and the oracle
reaches $0.825$, leaving a 6.3-point gap for an adaptive selector to
recover.

\Cref{tab:app-full-window-results} shows that some other windows offer
larger oracle gaps, but they correspond either to very early, noisier
prediction settings or to later windows where the downstream
intervention is less actionable. The main configuration sits near the
middle of the headroom range while preserving a plausible educational
intervention horizon.

\subsection{Oracle imitation improves, but not deployment accuracy}
\label{subsec:res-algo-vs-repr}

\begin{table}[t]
    \centering
    \small
    \setlength{\tabcolsep}{4pt}
    \caption{Main $(14\text{d},14\text{d})$ results. ``Best static ref.'' is the
    strongest single base model on that held-out test split, reported as a
    hindsight reference. The first block uses the 800-train/200-test regime
    that supports the BC/DQN/CQL mechanism comparison. The bottom block
    verifies the headline negative claim at $N{=}2{,}000$ test examples
    with an 8,000-buffer training set (full sweep in
    \Cref{tab:sample-size}). DQN here uses the canonical $\{0,1\}$
    oracle-match reward; the original log-probability shaping reward yields
    $0.158 \pm 0.025$ oracle agreement and $0.749 \pm 0.027$ test accuracy
    on the same seeds (Appendix~\ref{app:dqn-marginals}). CQL adds the
    standard conservative-$Q$ penalty to the same one-step bandit
    objective, swept across three penalty weights $\alpha$. BC-full minus
    best-static-reference accuracy is $-0.010$ (95\% CI $[-0.019,-0.001]$)
    in the small-$N$ regime and $0.000 \pm 0.001$ at $N{=}2{,}000$.
    Probs-only BC minus BC-full is $-0.004$ (CI $[-0.017,0.010]$).
    Disagreement-augmented BC minus BC-full is $-0.005$ (CI
    $[-0.014,0.004]$).}
    \floatlabel{tab:main-results}
    \begin{tabular}{lcc}
    \toprule
    Method & Oracle agreement & Test acc. \\
    \midrule
    \multicolumn{3}{l}{\emph{800-train / 200-test, 5 seeds}} \\
    Best static ref. & -- & $0.762 \pm 0.024$ \\
    Oracle & 1.000 & $0.825$ \\
    Held-out $10$-NN & $0.486 \pm 0.041$ & $0.747 \pm 0.037$ \\
    BC (full state) & $0.508 \pm 0.041$ & $0.752 \pm 0.026$ \\
    DQN (oracle-match) & $0.356 \pm 0.038$ & $0.743 \pm 0.030$ \\
    CQL ($\alpha{=}0.1$) & $0.497 \pm 0.054$ & $0.753 \pm 0.038$ \\
    CQL ($\alpha{=}1.0$) & $0.511 \pm 0.033$ & $0.750 \pm 0.041$ \\
    CQL ($\alpha{=}5.0$) & $0.505 \pm 0.044$ & $0.747 \pm 0.034$ \\
    BC (probs only) & $0.386 \pm 0.039$ & $0.748 \pm 0.018$ \\
    BC (+disagreement) & $0.516 \pm 0.039$ & $0.747 \pm 0.024$ \\
    \midrule
    \multicolumn{3}{l}{\emph{8,000-train / 2,000-test verification, 5 seeds}} \\
    Best static ref. (hindsight) & -- & $0.753 \pm 0.000$ \\
    BC (full state) & -- & $0.753 \pm 0.001$ \\
    BC (probs only) & -- & $0.747 \pm 0.001$ \\
    DQN (oracle-match) & -- & $0.738 \pm 0.003$ \\
    \bottomrule
    \end{tabular}
\end{table}

\begin{figure}[t]
    \centering
    \scriptsize
    \begin{tikzpicture}[x=20cm,y=0.27cm]
      \draw[->, thin] (0.695,0) -- (0.855,0);
      \foreach \x/\lab in {0.70/0.70,0.75/0.75,0.80/0.80,0.85/0.85} {
        \draw[thin] (\x,0.4) -- (\x,-0.4) node[below=1pt,font=\tiny] {\lab};
      }
      \draw[red!70, dashed, thin] (0.762,0.4) -- (0.762,9.6);
      \node[left,font=\tiny] at (0.694,1) {Oracle};
      \filldraw[blue!70!black] (0.825,1) circle (1.2pt);
      \node[left,font=\tiny] at (0.694,2) {Best static};
      \draw[red!70!black, line width=0.9pt] (0.738,2) -- (0.786,2);
      \filldraw[red!70!black] (0.762,2) circle (1.2pt);
      \node[left,font=\tiny] at (0.694,3) {BC (full)};
      \draw[black, line width=0.8pt] (0.726,3) -- (0.778,3);
      \filldraw[black] (0.752,3) circle (1.0pt);
      \node[left,font=\tiny] at (0.694,4) {CQL ($\alpha{=}1$)};
      \draw[black, line width=0.8pt] (0.709,4) -- (0.791,4);
      \filldraw[black] (0.750,4) circle (1.0pt);
      \node[left,font=\tiny] at (0.694,5) {BC (probs)};
      \draw[black, line width=0.8pt] (0.730,5) -- (0.766,5);
      \filldraw[black] (0.748,5) circle (1.0pt);
      \node[left,font=\tiny] at (0.694,6) {Held-out 10-NN};
      \draw[black, line width=0.8pt] (0.710,6) -- (0.784,6);
      \filldraw[black] (0.747,6) circle (1.0pt);
      \node[left,font=\tiny] at (0.694,7) {BC (+disagree)};
      \draw[black, line width=0.8pt] (0.723,7) -- (0.771,7);
      \filldraw[black] (0.747,7) circle (1.0pt);
      \node[left,font=\tiny] at (0.694,8) {CQL ($\alpha{=}5$)};
      \draw[black, line width=0.8pt] (0.713,8) -- (0.781,8);
      \filldraw[black] (0.747,8) circle (1.0pt);
      \node[left,font=\tiny] at (0.694,9) {DQN (oracle)};
      \draw[black, line width=0.8pt] (0.713,9) -- (0.773,9);
      \filldraw[black] (0.743,9) circle (1.0pt);
      \node[font=\tiny] at (0.775,-2.2) {Test accuracy (\Cref{tab:main-results}; static ref dashed)};
    \end{tikzpicture}
    \caption{Every learned selector clusters within $[0.74, 0.77]$ near
    the static reference. The oracle headroom of $0.063$ is not recovered.
    Bars: $\pm 1$ std over five seeds.}
    \floatlabel{fig:main-acc-bunching}
\end{figure}
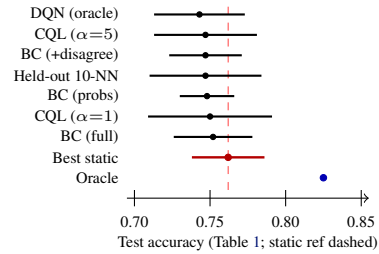

The local-consistency diagnostic at $(14\text{d},14\text{d})$ is
$0.388 \pm 0.010$. A held-out $10$-NN selector in the same standardized
state space reaches $0.486 \pm 0.041$ oracle agreement and
$0.747 \pm 0.037$ test accuracy, which calibrates local recoverability
from the current representation. BC, DQN, and CQL cluster between
$0.36$ and $0.51$ oracle agreement, above random ($1/K = 0.20$) but
below the $10$-NN ceiling. CQL matches BC's imitation ($0.51$ at
$\alpha{=}1.0$) while DQN reaches only $0.36$. The gap between them
isolates a missing pessimism term, since the supervisory signal is the
same. The log-probability shaping reward originally used for DQN
collapses oracle agreement below random because it is computed
relative to LR; Appendix~\ref{app:dqn-marginals} shows test accuracy
is roughly constant across reward variants ($0.743$ to $0.755$).

Yet none of the four learners exceeds the static reference of $0.762$
on held-out accuracy. Every method sits within $\pm 0.01$ of $0.748$
(\Cref{fig:main-acc-bunching}). The state ablations show the same
pattern: probabilities-only BC nearly matches the full state, and
disagreement-derived transforms improve oracle agreement slightly but
leave test accuracy unchanged. Of the paired intervals in
\Cref{tab:main-results}, only BC-full versus best static excludes
zero, and in the wrong direction; both representation comparisons span
zero. Richer offline encodings of the same base-model outputs do not
measurably improve deployment accuracy here. The 200-example test
split is not the cause: the bottom block of \Cref{tab:main-results}
shows that at $N{=}2{,}000$ held-out examples and an
$8{,}000$-example training buffer, BC's test accuracy
($0.753 \pm 0.001$) is statistically indistinguishable from the
hindsight static reference ($0.753 \pm 0.000$) and DQN remains below
both. Full 16-window results appear in
\Cref{tab:app-full-window-results}.

\subsection{The negative result is stable across windows and buffer sizes}
\label{subsec:res-window-stability}

\Cref{tab:app-full-window-results} shows that the
$(14\text{d},14\text{d})$ main case is not a one-off failure. Across
all 16 observation/prediction windows, BC without disagreement
augmentation trails the strongest single base model by 0.7 to 3.8
accuracy points, with a mean deficit of 1.6 points. The
disagreement-augmented selector also fails to close the gap: across
the same windows it trails that reference by 0.2 to 3.2 points, with a
mean deficit of 1.9 points.

The window sweep also clarifies what disagreement features change.
Augmentation improves oracle agreement in every window, by 0.4 to 11.9
points with a mean gain of 3.0 points; the corresponding held-out
accuracy change averages $-0.003$ and never exceeds $+0.009$.
Deterministic transforms of the probability vector increase oracle
agreement without producing a meaningful deployment benefit, which
supports the same representational diagnosis.

\subsubsection*{Sample-size sweep}

\begin{figure}[t]
  \centering
  \includegraphics[width=\columnwidth]{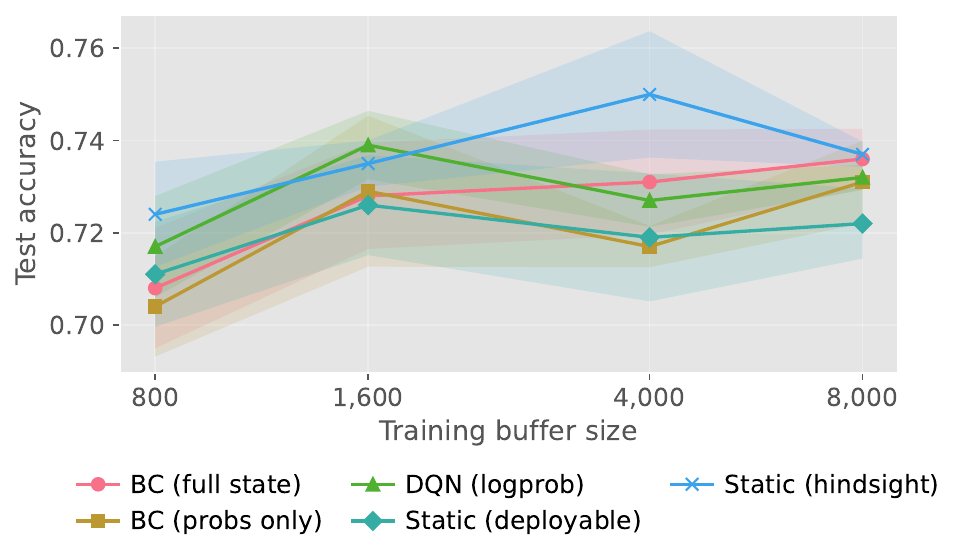}
  \caption{Sample-size sensitivity at $(14\text{d},14\text{d})$.
  Training buffer varied over $\{800, 1{,}600, 4{,}000, 8{,}000\}$ with
  the $20\%$ test split held fixed across rows. Bands: mean $\pm$ std
  over five seeds.}
  \floatlabel{fig:sample-size}
\end{figure}

\begin{table*}[t]
  \centering
  \caption{Sample-size sensitivity numbers underlying \Cref{fig:sample-size}.
  ``static-hind.'' is the strongest single base model on the held-out
  test split, picked with hindsight. ``static-deploy.'' is the base model
  selected on training accuracy and evaluated on the same test split, a
  deployable selector. The held-out test partition contains $2{,}000$
  examples and is fixed across rows.}
  \floatlabel{tab:sample-size}
  {\footnotesize\setlength{\tabcolsep}{3pt}\renewcommand{\arraystretch}{0.95}%
\begin{tabular}{r cc cc cc cc}
\toprule
buffer & $k$-NN full & $k$-NN probs & BC full acc & BC probs acc & DQN acc & $d_{\mathrm{TV}}$ & static-hind. & static-deploy. \\
\midrule
800 & $0.381\!\pm\!0.019$ & $0.500\!\pm\!0.023$ & $0.740\!\pm\!0.004$ & $0.737\!\pm\!0.007$ & $0.731\!\pm\!0.006$ & $0.046\!\pm\!0.018$ & $0.739\!\pm\!0.004$ & $0.734\!\pm\!0.003$ \\
1600 & $0.370\!\pm\!0.025$ & $0.511\!\pm\!0.022$ & $0.748\!\pm\!0.006$ & $0.744\!\pm\!0.005$ & $0.737\!\pm\!0.003$ & $0.045\!\pm\!0.025$ & $0.748\!\pm\!0.003$ & $0.739\!\pm\!0.001$ \\
4000 & $0.349\!\pm\!0.006$ & $0.519\!\pm\!0.007$ & $0.749\!\pm\!0.003$ & $0.745\!\pm\!0.004$ & $0.737\!\pm\!0.003$ & $0.027\!\pm\!0.010$ & $0.752\!\pm\!0.003$ & $0.738\!\pm\!0.006$ \\
8000 & $0.356\!\pm\!0.004$ & $0.540\!\pm\!0.002$ & $0.753\!\pm\!0.001$ & $0.747\!\pm\!0.001$ & $0.738\!\pm\!0.003$ & $0.028\!\pm\!0.010$ & $0.753\!\pm\!0.000$ & $0.744\!\pm\!0.002$ \\
\bottomrule
\end{tabular}
}
\end{table*}
\label{subsec:res-sample-size}

A natural reading of the main-text gap of $-0.010$ between BC and the
strongest single base model is that 200 test examples is too few to
distinguish the two. To test that explanation, we hold the test
partition fixed at $2{,}000$ examples and vary only the training-buffer
size. This
addresses both the sample-size concern and the test-stability concern
flagged by the diagnostic protocol's own design.
\Cref{tab:sample-size} reports the result.

The negative deployment claim survives at every buffer size. BC test
accuracy rises from $0.740 \pm 0.004$ at 800 training examples to
$0.753 \pm 0.001$ at $8{,}000$, but so does the hindsight static
reference, and the gap between them collapses to a mean difference of
roughly zero ($+0.000 \pm 0.001$ at $8{,}000$) without crossing into a
positive gain. Against the deployable static reference (the base model
picked from training-set accuracy), BC actually \emph{leads} by 0.6 to
0.9 points across all sizes. We treat that lead as suggestive because
it is small and depends on a deployment story whose details
(training-set picking rule, intervention budget, evaluation metric)
are specific to this task.

The sweep also clarifies the diagnostic story underneath that result.
Local $k$-NN consistency is roughly flat in the buffer size
($0.381 \to 0.356$ as the buffer grows by an order of magnitude),
consistent with the local-ambiguity diagnosis not being a small-sample
artifact. Buffer-to-test oracle shift, by contrast, decreases
monotonically ($d_{\mathrm{TV}} = 0.046 \to 0.028$), consistent with a
buffer that better approximates the test-time marginal as it grows.
Neither change moves deployment accuracy materially, which mirrors the
single-cache analysis in \Cref{subsec:res-dtv}.

\subsection{Hard-label tie-breaking does not drive the negative result}
\label{subsec:res-regret}

The hard-argmax oracle treats samples on which multiple base models are
simultaneously correct as if the choice between them mattered. The tie
rate (the buffer fraction with $\geq 2$ correct base models) is
$0.778 \pm 0.008$. We therefore also report mean per-sample regret: the
fraction of test samples on which the policy selects an incorrect model
when at least one correct model exists (the oracle attains $0$).
\begin{center}
  \small
  \begin{tabular}{lcc}
  \toprule
  Method & Oracle agree. & Mean regret \\
  \midrule
  Best static ref.       & --                & $0.059$           \\
  BC (full state)        & $0.508 \pm 0.041$ & $0.089 \pm 0.016$ \\
  DQN (oracle-match)     & $0.356 \pm 0.038$ & $0.101 \pm 0.009$ \\
  CQL ($\alpha{=}1.0$)   & $0.511 \pm 0.033$ & $0.094 \pm 0.012$ \\
  \bottomrule
  \end{tabular}
\end{center}
Two things follow. First, oracle agreement spans $0.36$ to $0.52$
while regret clusters tightly in $[0.089, 0.101]$, so hard-label
agreement overstates how much BC, DQN, and CQL differ on a
deployment-relevant metric. Second, the strongest base model still
leads in regret ($0.059$), so the negative result is not a
tie-breaking artifact. The high tie rate does qualify Stage 1: with
$0.778$ of buffer samples having $\geq 2$ correct base models, the
$k$-NN consistency of $0.388$ partly reflects arg-max noise among
near-tied probabilities, and is best read as an upper bound on the
signal recoverable by any hard-label local rule.

\subsection{Buffer-to-test oracle shift is locally non-trivial}
\label{subsec:res-dtv}

\begin{figure}[t]
  \centering
  \includegraphics[width=\columnwidth]{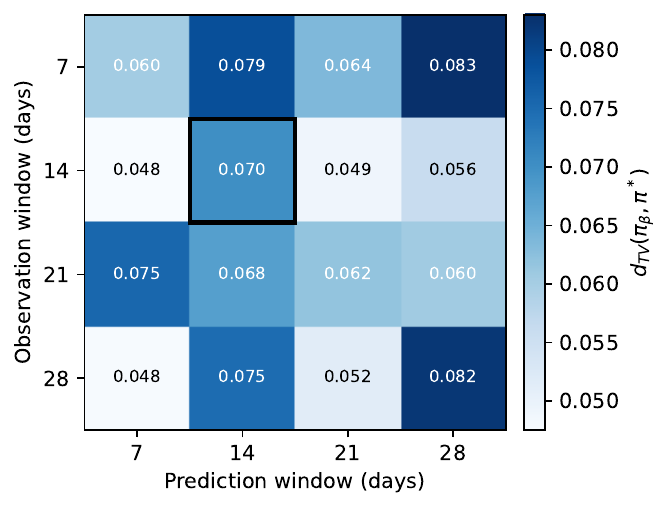}
  \caption{Buffer-to-test marginal $d_{\mathrm{TV}}(\pi_\beta,\pi^\star)$
  across the 16 windows. All values are positive but modest. Main
  configuration: $0.070 \pm 0.015$. Full table and controlled-classifier
  analysis in \Cref{tab:app-dist-shift}.}
  \floatlabel{fig:oracle-shift}
\end{figure}

\Cref{fig:oracle-shift} shows that the oracle-label distribution
induced by the offline buffer is not identical to the test-time
oracle. Across all 16 window configurations,
$d_{\mathrm{TV}}(\pi_\beta,\pi^\star)$ ranges from $0.048$ to
$0.083$, with a mean of $0.063 \pm 0.011$. On the main configuration
it is $0.070 \pm 0.015$.

The controlled classifier analysis summarized in
\Cref{tab:app-dist-shift} reaches the same conclusion. On
$(14\text{d},14\text{d})$, a classifier trained to imitate the buffer
oracle is closer to the buffer than to the test oracle by
$+0.016 \pm 0.040$, confirming that shift is in the right direction to
hurt deployment. A separate robustness check using external
stratifiers is reported in Appendix~\ref{app:external-stratifier}.

The marginal $d_{\mathrm{TV}}$ in \Cref{eq:dtv} averages over the
state, so a small marginal can hide substantial pointwise shift. We
therefore also estimate the conditional shift
$\mathbb{E}_s[d_{\mathrm{TV}}(\pi_\beta(\cdot\mid s),
\pi^\star(\cdot\mid s))]$ \label{eq:cond-dtv} by approximating each
per-state distribution from its $k{=}10$ nearest neighbors (buffer-side
for $\pi_\beta$, test-side for $\pi^\star$). On the main configuration
the conditional $d_{\mathrm{TV}}$ averages $0.288 \pm 0.024$, roughly
four times the marginal, with $95$th percentile $0.521 \pm 0.040$ and
worst-state $0.860 \pm 0.049$. Pointwise shift is therefore much
larger than the marginal suggests, with a small tail of states
approaching full disagreement.

If local shift were the dominant failure mode, CQL's conservatism term
(which explicitly tightens toward $\pi_\beta$) should hurt deployment
more than DQN's unregularized objective. It does not. BC, DQN, and CQL
differ sharply in oracle imitation ($0.36$ to $0.51$) but converge on
test accuracy ($0.743$ to $0.753$, within $0.01$ across all three CQL
penalty weights). Local support mismatch is substantial and works
against deployment, but it is not the main driver of the result. The
offline state simply does not identify a deployment-winning action
reliably enough, and that limitation is shared across learners
regardless of how aggressively they regularize toward the buffer.

\section{Discussion: An Offline Diagnostic Procedure}
\label{sec:discussion}

The combined diagnostic procedure points to a locally ambiguous state
representation as the primary failure mode. Held-out accuracy is nearly unmoved
by disagreement-derived feature transforms, by adding the canonical
offline-RL conservatism term, or by enlarging the buffer tenfold.
Buffer-to-test oracle shift moves it only marginally, even when the
shift is measured per-state.

\subsection{Why offline-RL constraints may not close this gap}
\label{subsec:disc-cql-iql}

Conservative offline-RL methods (CQL~\cite{kumar:cql},
IQL~\cite{kostrikov:iql}, BRAC~\cite{wu:brac}, and
BEAR~\cite{kumar:bear}) target $Q$-value
overestimation at unseen $(s,a)$ under bootstrap. This task exhibits a
different failure mode. Setting $\gamma{=}0$ eliminates bootstrap by
construction, and the ambiguity is in the $(s, a^*)$ labelling itself.
Similar states disagree on the oracle action $61.2\%$ of the time.
Conservatism reweights the decision rule under the existing labels;
the labels themselves are unchanged. It therefore cannot recover
signal that the $(s, a^\star)$ pairs do not already contain.

\Cref{tab:main-results} bears this out. Adding the CQL penalty to the
same one-step objective recovers oracle agreement of $0.51$ at
$\alpha{=}1.0$, comparable to BC and well above DQN's $0.36$. Across
$\alpha \in \{0.1, 1.0, 5.0\}$, test accuracy moves by less than a
point and never reaches the static reference of $0.762$. Conservatism
closes the imitation gap without moving deployment, even though the
conditional shift $\mathbb{E}_s[d_{\mathrm{TV}}] \approx 0.29$
(\Cref{eq:cond-dtv}) is substantial: the gap to the static reference
appears to be limited by the representation rather than by distance to
$\pi_\beta$.

\subsection{Using the diagnostics and benchmark implications}
\label{subsec:disc-practical}

As a pre-deployment checklist, the procedure first asks whether oracle
headroom is large enough to justify any adaptive selector. If headroom
exists, it asks whether a simple local predictor can recover oracle
actions from the current state. When both $k$-NN consistency and
held-out $10$-NN accuracy are low, additional offline-RL tuning is
unlikely to pay off; the next iteration should focus on state design,
relabeling, or data collection. Heavier offline learners or
offline-to-online adaptation~\cite{mandel:ope} are worth the cost only
when local recoverability is already meaningful. The same four
quantities (oracle gap, $k$-NN consistency, a held-out local selector,
and a buffer-to-test oracle-shift measure) make a negative
offline-selector result interpretable in benchmark settings.

\subsection{Fairness and intervention-cost considerations}
\label{subsec:disc-fairness}

The inclusion criterion ($\geq 10$ events, $\geq 3$ active days,
$\geq 28$-day activity span) excludes roughly $85\%$ of enrolled
learners, mostly casual browsers, so our claims are conditioned on the
moderately engaged subpopulation. A learned selector also triggers
downstream human intervention, so selection errors that systematically
favor one base model in one subpopulation may translate into
inequitable intervention coverage. An offline meta-learner whose
failure modes are unmapped across demographic strata is not yet
suitable for intervention-triggering use.

\section{Limitations and Conclusion}
\label{sec:conclusion}

All experiments draw from one introductory computer science curriculum
at one institution; the diagnostics should be recomputed on any new
dataset. The inclusion criteria bias the sample toward moderately
engaged learners, and the contextual-bandit reduction ($\gamma{=}0$)
cannot model within-course state transitions. The offline setting also
precludes the online exploration our diagnosis motivates, so we can
diagnose offline limits but cannot measure what online adaptation would
achieve here.

On this task, the diagnostic procedure points to local label ambiguity
as the primary offline bottleneck. Adding conservatism does not close
the gap in our experiments, and accounting for buffer-to-test shift
does not either. The combined procedure is a lightweight pre-deployment
checklist for deciding whether the offline state is informative enough
to support model selection: when it flags low consistency and shared
BC/DQN/CQL failure, the next iteration is more likely to pay off by
changing the data than by changing the learner.


\bibliographystyle{icml2026}
\bibliography{main}

@article{levine:offline,
  title  = {Offline reinforcement learning: Tutorial, review, and perspectives on open problems},
  author = {Levine, Sergey and Kumar, Aviral and Tucker, George and Fu, Justin},
  journal = {arXiv preprint arXiv:2005.01643},
  year   = {2020},
  doi    = {10.48550/arXiv.2005.01643},
  url    = {https://arxiv.org/abs/2005.01643}
}

@article{rice:selection,
  title     = {The algorithm selection problem},
  author    = {Rice, John R.},
  journal   = {Advances in Computers},
  volume    = {15},
  pages     = {65--118},
  year      = {1976},
  publisher = {Elsevier},
  doi       = {10.1016/S0065-2458(08)60520-3},
  url       = {https://doi.org/10.1016/S0065-2458(08)60520-3}
}

@article{cruz:des,
  title   = {Dynamic classifier selection: Recent advances and perspectives},
  author  = {Cruz, Rafael M. O. and Sabourin, Robert and Cavalcanti, George D. C.},
  journal = {Information Fusion},
  volume  = {41},
  pages   = {195--216},
  year    = {2018},
  doi     = {10.1016/j.inffus.2017.09.010},
  url     = {https://doi.org/10.1016/j.inffus.2017.09.010}
}

@article{ko:des,
  title   = {From dynamic classifier selection to dynamic ensemble selection},
  author  = {Ko, Albert Hung-Ren and Sabourin, Robert and Britto, Alceu de Souza},
  journal = {Pattern Recognition},
  volume  = {41},
  number  = {5},
  pages   = {1718--1731},
  year    = {2008},
  doi     = {10.1016/j.patcog.2007.10.015},
  url     = {https://doi.org/10.1016/j.patcog.2007.10.015}
}

@article{cruz:metades,
  title   = {{META-DES}: {A} dynamic ensemble selection framework using meta-learning},
  author  = {Cruz, Rafael M. O. and Sabourin, Robert and Cavalcanti, George D. C. and Ren, Tsang Ing},
  journal = {Pattern Recognition},
  volume  = {48},
  number  = {5},
  pages   = {1925--1935},
  year    = {2015},
  doi     = {10.1016/j.patcog.2014.12.003},
  url     = {https://doi.org/10.1016/j.patcog.2014.12.003}
}

@article{liu:clickstream,
  title   = {Predicting Student Performance Using Clickstream Data and Machine Learning},
  author  = {Liu, Yutong and Fan, Si and Xu, Shuxiang and Sajjanhar, Atul and Yeom, Soonja and Wei, Yuchen},
  journal = {Education Sciences},
  volume  = {13},
  number  = {1},
  pages   = {17},
  year    = {2023},
  doi     = {10.3390/educsci13010017},
  url     = {https://doi.org/10.3390/educsci13010017}
}

@article{taylor:stopout,
  title   = {Likely to Stop? Predicting Stopout in Massive Open Online Courses},
  author  = {Taylor, Colin and Veeramachaneni, Kalyan and O'Reilly, Una-May},
  journal = {arXiv preprint arXiv:1408.3382},
  year    = {2014},
  doi     = {10.48550/arXiv.1408.3382},
  url     = {https://arxiv.org/abs/1408.3382}
}

@inproceedings{kumar:cql,
  title     = {Conservative Q-Learning for Offline Reinforcement Learning},
  author    = {Kumar, Aviral and Zhou, Aurick and Tucker, George and Levine, Sergey},
  booktitle = {Advances in Neural Information Processing Systems (NeurIPS)},
  volume    = {33},
  pages     = {1179--1191},
  year      = {2020},
  url       = {https://papers.nips.cc/paper/2020/hash/0d2b2061826a5df3221116a5085a6052-Abstract.html}
}

@article{mnih:dqn,
  title   = {Playing Atari with Deep Reinforcement Learning},
  author  = {Mnih, Volodymyr and Kavukcuoglu, Koray and Silver, David and Graves, Alex and Antonoglou, Ioannis and Wierstra, Daan and Riedmiller, Martin},
  journal = {arXiv preprint arXiv:1312.5602},
  year    = {2013},
  doi     = {10.48550/arXiv.1312.5602},
  url     = {https://arxiv.org/abs/1312.5602}
}

@article{kostrikov:iql,
  title   = {Offline Reinforcement Learning with Implicit {Q}-Learning},
  author  = {Kostrikov, Ilya and Nair, Ashvin and Levine, Sergey},
  journal = {arXiv preprint arXiv:2110.06169},
  year    = {2021},
  url     = {https://arxiv.org/abs/2110.06169}
}

@inproceedings{fujimoto:bcq,
  title     = {Off-Policy Deep Reinforcement Learning without Exploration},
  author    = {Fujimoto, Scott and Meger, David and Precup, Doina},
  booktitle = {Proceedings of the 36th International Conference on Machine Learning},
  series    = {Proceedings of Machine Learning Research},
  volume    = {97},
  pages     = {2052--2062},
  year      = {2019},
  publisher = {PMLR},
  url       = {https://proceedings.mlr.press/v97/fujimoto19a.html}
}

@inproceedings{kumar:bear,
  title     = {Stabilizing Off-Policy Q-Learning via Bootstrapping Error Reduction},
  author    = {Kumar, Aviral and Fu, Justin and Soh, Matthew and Tucker, George and Levine, Sergey},
  booktitle = {Advances in Neural Information Processing Systems (NeurIPS)},
  volume    = {32},
  pages     = {11761--11771},
  year      = {2019},
  url       = {https://papers.neurips.cc/paper/2019/hash/c2073ffa77b5357a498057413bb09d3a-Abstract.html}
}

@article{wu:brac,
  title   = {Behavior Regularized Offline Reinforcement Learning},
  author  = {Wu, Yifan and Tucker, George and Nachum, Ofir},
  journal = {arXiv preprint arXiv:1911.11361},
  year    = {2019},
  url     = {https://arxiv.org/abs/1911.11361}
}

@article{wolpert:stacking,
  title   = {Stacked generalization},
  author  = {Wolpert, David H.},
  journal = {Neural Networks},
  volume  = {5},
  number  = {2},
  pages   = {241--259},
  year    = {1992},
  doi     = {10.1016/S0893-6080(05)80023-1},
  url     = {https://doi.org/10.1016/S0893-6080(05)80023-1}
}

@article{casado:metalearning,
  title   = {Using Meta-Learning to predict student performance in virtual learning environments},
  author  = {Casado Hidalgo, {\'A}ngel and Moreno-Ger, Pablo and de la Fuente-Valent{\'\i}n, Luis},
  journal = {Applied Intelligence},
  volume  = {52},
  number  = {3},
  pages   = {3352--3365},
  year    = {2022},
  doi     = {10.1007/s10489-021-02613-x},
  url     = {https://doi.org/10.1007/s10489-021-02613-x}
}

@inproceedings{dalipi:review,
  title     = {{MOOC} dropout prediction using machine learning techniques: {Review} and research challenges},
  author    = {Dalipi, Fisnik and Imran, Ali Shariq and Kastrati, Zenun},
  booktitle = {2018 IEEE Global Engineering Education Conference (EDUCON)},
  pages     = {1007--1014},
  year      = {2018},
  doi       = {10.1109/EDUCON.2018.8363340},
  url       = {https://ieeexplore.ieee.org/document/8363340}
}

@inproceedings{berens:early,
  title     = {Early Prediction of Student Dropout in Higher Education using Machine Learning Models},
  author    = {Goren, Or and Cohen, Liron and Rubinstein, Amir},
  booktitle = {Proceedings of the 17th International Conference on Educational Data Mining (EDM)},
  pages     = {349--359},
  year      = {2024},
  url       = {https://educationaldatamining.org/edm2024/proceedings/2024.EDM-short-papers.32/index.html}
}

@inproceedings{lan:bandits,
  title     = {A Contextual Bandits Framework for Personalized Learning Action Selection},
  author    = {Lan, Andrew S. and Baraniuk, Richard G.},
  booktitle = {Proceedings of the 9th International Conference on Educational Data Mining (EDM)},
  pages     = {424--429},
  year      = {2016},
  url       = {https://educationaldatamining.org/EDM2016/proceedings/paper_63.pdf}
}

@inproceedings{mandel:ope,
  title     = {Offline Policy Evaluation Across Representations with Applications to Educational Games},
  author    = {Mandel, Travis and Liu, Yun-En and Levine, Sergey and Brunskill, Emma and Popovi{\'c}, Zoran},
  booktitle = {Proceedings of the 13th International Conference on Autonomous Agents and Multiagent Systems (AAMAS)},
  pages     = {1077--1084},
  year      = {2014},
  publisher = {International Foundation for Autonomous Agents and Multiagent Systems},
  url       = {https://www.ifaamas.org/Proceedings/aamas2014/aamas/p1077.pdf}
}

@inproceedings{dudik:doublyrobust,
  title     = {Doubly Robust Policy Evaluation and Learning},
  author    = {Dud{\'i}k, Miroslav and Langford, John and Li, Lihong},
  booktitle = {Proceedings of the 28th International Conference on Machine Learning (ICML)},
  pages     = {1097--1104},
  year      = {2011},
  publisher = {Omnipress},
  url       = {https://icml.cc/2011/papers/554_icmlpaper.pdf}
}

@inproceedings{swaminathan:crm,
  title     = {Counterfactual Risk Minimization: Learning from Logged Bandit Feedback},
  author    = {Swaminathan, Adith and Joachims, Thorsten},
  booktitle = {Proceedings of the 32nd International Conference on Machine Learning},
  series    = {Proceedings of Machine Learning Research},
  volume    = {37},
  pages     = {814--823},
  year      = {2015},
  publisher = {PMLR},
  url       = {https://proceedings.mlr.press/v37/swaminathan15.html}
}

@article{salleh:randomforest,
  title   = {Predicting Student Dropout in Self-Paced {MOOC} Course Using Random Forest Model},
  author  = {Dass, Sheran and Gary, Kevin and Cunningham, James},
  journal = {Information},
  volume  = {12},
  number  = {11},
  pages   = {476},
  year    = {2021},
  doi     = {10.3390/info12110476},
  url     = {https://doi.org/10.3390/info12110476}
}

@article{nagrecha:temporal,
  title   = {Temporal predication of dropouts in {MOOCs}: Reaching the low hanging fruit through stacking generalization},
  author  = {Xing, Wanli and Chen, Xin and Stein, Jared and Marcinkowski, Michael},
  journal = {Computers in Human Behavior},
  volume  = {58},
  pages   = {119--129},
  year    = {2016},
  doi     = {10.1016/j.chb.2015.12.007},
  url     = {https://doi.org/10.1016/j.chb.2015.12.007}
}

@article{lykourentzou:ensemble,
  title   = {Dropout prediction in e-learning courses through the combination of machine learning techniques},
  author  = {Lykourentzou, Ioanna and Giannoukos, Ioannis and Nikolopoulos, Vassilis and Mpardis, George and Loumos, Vassili},
  journal = {Computers \& Education},
  volume  = {53},
  number  = {3},
  pages   = {950--965},
  year    = {2009},
  doi     = {10.1016/j.compedu.2009.05.010},
  url     = {https://doi.org/10.1016/j.compedu.2009.05.010}
}

\clearpage
\appendix
\onecolumn
\setlength{\textfloatsep}{10pt plus 2pt minus 4pt}
\setlength{\floatsep}{8pt plus 2pt minus 2pt}
\setlength{\intextsep}{8pt plus 2pt minus 2pt}

\section{Feature and Setup Details}
\label{app:feature-setup}

The filtered edX corpus contains $34{,}303$ qualifying student-course pairs
from $21{,}990$ students after requiring at least 10 events, at least 3
active days, and at least 28 days of activity span. These filters bias the
analysis toward moderately engaged learners, but they ensure that every
sample supports the observation/prediction windows evaluated in the main
paper.

From each observation window we derive 28 engineered behavioral features in
six categories. These complement the modality flag and base-model
probability vector used by the selector. \Cref{tab:app-feature-categories}
summarizes the taxonomy used throughout the paper.

\begin{table}[H]
\centering
\small
\caption{Feature categories with representative examples. All 28 features
enter the selector state as standardized scalars.}
\floatlabel{tab:app-feature-categories}
\begin{tabular}{l l c}
\toprule
\textbf{Category} & \textbf{Examples} & \textbf{Count} \\
\midrule
Volume         & total\_events, active\_days                 & 4 \\
Content ratios & video\_ratio, problem\_ratio                & 9 \\
Consistency    & event\_variance, max\_gap\_days             & 4 \\
Temporal       & first/last-week ratio, days\_since\_last    & 4 \\
Trends         & engagement\_slope, peak\_to\_avg            & 2 \\
Advanced       & streak\_length, decay\_rate, Gini           & 5 \\
\midrule
\textbf{Total} &                                             & \textbf{28} \\
\bottomrule
\end{tabular}
\end{table}

\section{Main-Case Diagnostics and Raw Counts}
\label{app:main-case}

The body's \Cref{fig:main-acc-bunching} compresses the main
$(14\text{d},14\text{d})$ test-accuracy comparison into one view. BC,
DQN, CQL, and held-out $10$-NN all remain near the static reference of
$0.762$ while the per-instance oracle reaches $0.825$. Each seed
evaluates exactly 200 held-out examples, so the $0.010$ accuracy gap
between BC-full and the strongest single base model corresponds to
roughly two test examples per seed. That scale is easy to lose in the
mean-based summaries reported in the body. The corresponding
buffer/test counts and main-case action marginals appear in
\Cref{tab:app-main-case-marginals}.

\begin{center}
\small
\captionsetup{hypcap=false}
\captionof{table}{Main $(14\text{d},14\text{d})$ action marginals averaged over 5
seeds. Each seed uses an 800-example CV buffer and a 200-example held-out
test split. The same regime yields $k$-NN consistency
$0.388 \pm 0.010$, BC buffer-oracle agreement $0.541 \pm 0.012$,
BC test-oracle agreement $0.508 \pm 0.041$,
$d_{\mathrm{TV}}(\pi_\beta,\pi^\star)=0.070 \pm 0.015$, and a
classifier-to-buffer minus classifier-to-test distance of
$+0.016 \pm 0.040$.}
\floatlabel{tab:app-main-case-marginals}
\begin{tabular}{lccccc}
\toprule
\textbf{Action source} & \textbf{LR} & \textbf{RF} & \textbf{GB} & \textbf{CalRF} & \textbf{Stack} \\
\midrule
Buffer oracle $\pi_\beta$ & 0.271 & 0.182 & 0.331 & 0.128 & 0.088 \\
Test oracle $\pi^\star$ & 0.265 & 0.217 & 0.321 & 0.111 & 0.086 \\
Controlled clf $\pi_{\mathrm{clf}}$ & 0.343 & 0.106 & 0.487 & 0.052 & 0.012 \\
\bottomrule
\end{tabular}
\end{center}

\section{DQN Action-Marginal Diagnosis}
\label{app:dqn-marginals}

The body \Cref{tab:main-results} reports DQN under the canonical
$\{0,1\}$ oracle-match reward, which attains
$0.356 \pm 0.038$ oracle agreement on the $(14\text{d},14\text{d})$
configuration. The log-probability shaping reward used in earlier
drafts attains only $0.158 \pm 0.025$ oracle agreement on the same
seeds, below the $1/K = 0.20$ uniform-random baseline on $K{=}5$
actions. This appendix explains why, by holding the buffer, base
models, CV partition, and DQN architecture fixed and sweeping only the
reward shaping function defined in \Cref{sec:cascade}. Five reward
variants are evaluated on the same $(14\text{d},14\text{d})$ buffer
with five seeds: $\mathrm{logprob}$ (the original variant),
$\mathrm{clipped}$ correctness, $\mathrm{oracle\_match}$ (binary
correctness on the chosen base model, used in the body),
$\mathrm{margin}$, and a $\mathrm{hybrid}$ correctness-plus-margin
combination.

\begin{table}[H]
\centering
\caption{DQN action-marginal sweep on $(14\text{d},14\text{d})$, five seeds.
``oracle agree'' is the fraction of test samples on which the DQN's chosen
action equals the test oracle's. ``top share'' is the largest DQN
selected-action marginal. The five right-most columns are the DQN's
selected-action marginal on the test set, one column per base model.
The ``oracle (ref)'' row reports the test-set oracle's marginal as a
reference target.}
\floatlabel{tab:app-dqn-marginals}
{\footnotesize\setlength{\tabcolsep}{3pt}\renewcommand{\arraystretch}{0.95}%
\begin{tabular}{l ccc ccccc}
\toprule
reward & oracle agree & test acc & top share & LR & RF & GB & CalRF & Stack \\
\midrule
\textit{oracle (ref)} & 1.000 & --- & --- & 0.270 & 0.200 & 0.321 & 0.122 & 0.087 \\
\midrule
logprob & $0.158\!\pm\!0.026$ & $0.749\!\pm\!0.029$ & $0.422\!\pm\!0.103$ & 0.000 & 0.153 & 0.239 & 0.394 & 0.214 \\
clipped & $0.165\!\pm\!0.046$ & $0.749\!\pm\!0.023$ & $0.321\!\pm\!0.052$ & 0.077 & 0.305 & 0.147 & 0.207 & 0.264 \\
oracle\_match & $0.356\!\pm\!0.037$ & $0.743\!\pm\!0.033$ & $0.364\!\pm\!0.054$ & 0.232 & 0.301 & 0.279 & 0.107 & 0.081 \\
margin & $0.248\!\pm\!0.040$ & $0.750\!\pm\!0.025$ & $0.561\!\pm\!0.091$ & 0.561 & 0.093 & 0.051 & 0.054 & 0.241 \\
hybrid & $0.349\!\pm\!0.052$ & $0.755\!\pm\!0.031$ & $0.361\!\pm\!0.052$ & 0.329 & 0.228 & 0.243 & 0.081 & 0.119 \\
\bottomrule
\end{tabular}
}
\end{table}

Two patterns explain the headline number. First, the
$\mathrm{logprob}$ reward used in earlier drafts is computed relative
to the LR baseline,
$R_{\mathrm{log}}(s, a) = \log p_a^{(y)}(s) - \log p_{\mathrm{LR}}^{(y)}(s)$.
On samples where LR is itself the oracle action, the chosen and
baseline log-probabilities cancel and the buffer reward for that
sample is exactly zero. On samples where another model dominates LR,
the reward is positive. The DQN therefore observes that selecting LR is
never net-positive in the buffer, and across all $1{,}000$ test
predictions in the five-seed sweep it \emph{never selects LR}, despite
the test oracle picking LR 270 times. The LR-marginal cell in
\Cref{tab:app-dqn-marginals} for the $\mathrm{logprob}$ row is exactly
$0.000$. The $0.158$ logprob oracle agreement is dominated by these
unrecoverable LR samples.

Second, the failure is reward-specific. The $\mathrm{oracle\_match}$
reward (binary correctness, no LR-relative shaping) recovers an LR
marginal of $0.232$, close to the oracle's $0.270$, and lifts oracle
agreement from $0.158 \pm 0.026$ to $0.356 \pm 0.037$. The
$\mathrm{hybrid}$ variant reaches $0.349 \pm 0.052$. Importantly,
deployment-relevant test accuracy is roughly constant across reward
variants. The five rows span $0.743$ to $0.755$, all within their
mutual confidence intervals and all below the strongest single base
model reference of $0.762 \pm 0.024$. The choice of reward shaping
rearranges which samples the DQN gets right at the action level without
materially shifting the deployment metric.

This decomposition tightens the Stage-2 reading in
\Cref{subsec:res-algo-vs-repr}. The original logic (``BC and DQN fail
by similar margins, indicating a shared bottleneck above both
algorithms'') was strained when oracle agreements differed by $0.508$
(BC) vs.\ $0.158$ (logprob DQN). With the reward sweep, the
shared-failure claim is supported on the test-accuracy axis, where
every reward and every comparator lands near $0.75$, below the $0.762$
static reference. It is qualified on the oracle-agreement axis, where
reward shaping accounts for most of the original BC--DQN gap, and the
body's $\mathrm{oracle\_match}$ row at $0.356 \pm 0.038$ already lifts
the comparison above random. The representational diagnosis still
holds, and the algorithmic-versus-representational separation is now
sharper because the reward-shaping confound is identified.

\section{Full 16-Window Aggregated Results}
\label{app:full-window-results}

\Cref{tab:app-full-window-results} gives the full 16-window summary
underlying the body claims about local consistency, behavioral cloning, and
disagreement augmentation.

\begin{table}[H]
\centering
\caption{Full 16-window summary of local $k$-NN consistency, BC oracle
agreement, BC held-out accuracy, and the best single-model reference, with
and without disagreement augmentation.}
\floatlabel{tab:app-full-window-results}
{\footnotesize\setlength{\tabcolsep}{3pt}\renewcommand{\arraystretch}{0.95}%
\begin{tabular}{ccl cccc}
\toprule
obs & pred & augment & $k$-NN cons. & BC agree & BC acc & best single \\
\midrule
7 & 7 & disagreement & $0.395\!\pm\!0.036$ & $0.525\!\pm\!0.040$ & $0.727\!\pm\!0.017$ & $0.739\!\pm\!0.015$ \\
7 & 7 & none & $0.394\!\pm\!0.035$ & $0.406\!\pm\!0.044$ & $0.730\!\pm\!0.020$ & $0.739\!\pm\!0.015$ \\
7 & 14 & disagreement & $0.375\!\pm\!0.019$ & $0.439\!\pm\!0.043$ & $0.708\!\pm\!0.020$ & $0.731\!\pm\!0.021$ \\
7 & 14 & none & $0.374\!\pm\!0.020$ & $0.413\!\pm\!0.043$ & $0.722\!\pm\!0.019$ & $0.731\!\pm\!0.021$ \\
7 & 21 & disagreement & $0.397\!\pm\!0.021$ & $0.478\!\pm\!0.044$ & $0.744\!\pm\!0.015$ & $0.765\!\pm\!0.012$ \\
7 & 21 & none & $0.395\!\pm\!0.021$ & $0.472\!\pm\!0.039$ & $0.758\!\pm\!0.008$ & $0.765\!\pm\!0.012$ \\
7 & 28 & disagreement & $0.442\!\pm\!0.023$ & $0.549\!\pm\!0.029$ & $0.819\!\pm\!0.011$ & $0.836\!\pm\!0.014$ \\
7 & 28 & none & $0.440\!\pm\!0.022$ & $0.539\!\pm\!0.032$ & $0.822\!\pm\!0.017$ & $0.836\!\pm\!0.014$ \\
14 & 7 & disagreement & $0.393\!\pm\!0.014$ & $0.475\!\pm\!0.023$ & $0.740\!\pm\!0.020$ & $0.758\!\pm\!0.016$ \\
14 & 7 & none & $0.392\!\pm\!0.013$ & $0.419\!\pm\!0.048$ & $0.744\!\pm\!0.007$ & $0.758\!\pm\!0.016$ \\
14 & 14 & disagreement & $0.389\!\pm\!0.010$ & $0.516\!\pm\!0.039$ & $0.747\!\pm\!0.024$ & $0.762\!\pm\!0.024$ \\
14 & 14 & none & $0.388\!\pm\!0.010$ & $0.508\!\pm\!0.041$ & $0.752\!\pm\!0.026$ & $0.762\!\pm\!0.024$ \\
14 & 21 & disagreement & $0.377\!\pm\!0.015$ & $0.522\!\pm\!0.033$ & $0.745\!\pm\!0.031$ & $0.747\!\pm\!0.022$ \\
14 & 21 & none & $0.377\!\pm\!0.015$ & $0.487\!\pm\!0.043$ & $0.736\!\pm\!0.026$ & $0.747\!\pm\!0.022$ \\
14 & 28 & disagreement & $0.409\!\pm\!0.013$ & $0.532\!\pm\!0.036$ & $0.754\!\pm\!0.022$ & $0.786\!\pm\!0.013$ \\
14 & 28 & none & $0.409\!\pm\!0.013$ & $0.512\!\pm\!0.031$ & $0.753\!\pm\!0.014$ & $0.786\!\pm\!0.013$ \\
21 & 7 & disagreement & $0.371\!\pm\!0.013$ & $0.481\!\pm\!0.026$ & $0.720\!\pm\!0.021$ & $0.730\!\pm\!0.018$ \\
21 & 7 & none & $0.371\!\pm\!0.012$ & $0.430\!\pm\!0.016$ & $0.718\!\pm\!0.018$ & $0.730\!\pm\!0.018$ \\
21 & 14 & disagreement & $0.365\!\pm\!0.011$ & $0.468\!\pm\!0.039$ & $0.740\!\pm\!0.009$ & $0.755\!\pm\!0.012$ \\
21 & 14 & none & $0.364\!\pm\!0.010$ & $0.464\!\pm\!0.030$ & $0.747\!\pm\!0.009$ & $0.755\!\pm\!0.012$ \\
21 & 21 & disagreement & $0.377\!\pm\!0.017$ & $0.499\!\pm\!0.040$ & $0.721\!\pm\!0.022$ & $0.751\!\pm\!0.018$ \\
21 & 21 & none & $0.375\!\pm\!0.018$ & $0.482\!\pm\!0.044$ & $0.728\!\pm\!0.028$ & $0.751\!\pm\!0.018$ \\
21 & 28 & disagreement & $0.367\!\pm\!0.012$ & $0.443\!\pm\!0.016$ & $0.713\!\pm\!0.034$ & $0.742\!\pm\!0.004$ \\
21 & 28 & none & $0.365\!\pm\!0.013$ & $0.412\!\pm\!0.010$ & $0.704\!\pm\!0.020$ & $0.742\!\pm\!0.004$ \\
28 & 7 & disagreement & $0.390\!\pm\!0.015$ & $0.472\!\pm\!0.029$ & $0.766\!\pm\!0.026$ & $0.788\!\pm\!0.022$ \\
28 & 7 & none & $0.389\!\pm\!0.015$ & $0.451\!\pm\!0.022$ & $0.777\!\pm\!0.025$ & $0.788\!\pm\!0.022$ \\
28 & 14 & disagreement & $0.393\!\pm\!0.014$ & $0.484\!\pm\!0.078$ & $0.755\!\pm\!0.023$ & $0.777\!\pm\!0.022$ \\
28 & 14 & none & $0.392\!\pm\!0.015$ & $0.465\!\pm\!0.056$ & $0.758\!\pm\!0.021$ & $0.777\!\pm\!0.022$ \\
28 & 21 & disagreement & $0.350\!\pm\!0.013$ & $0.463\!\pm\!0.029$ & $0.727\!\pm\!0.020$ & $0.740\!\pm\!0.008$ \\
28 & 21 & none & $0.349\!\pm\!0.014$ & $0.448\!\pm\!0.047$ & $0.729\!\pm\!0.023$ & $0.740\!\pm\!0.008$ \\
28 & 28 & disagreement & $0.383\!\pm\!0.023$ & $0.506\!\pm\!0.049$ & $0.742\!\pm\!0.024$ & $0.757\!\pm\!0.019$ \\
28 & 28 & none & $0.381\!\pm\!0.022$ & $0.469\!\pm\!0.045$ & $0.736\!\pm\!0.035$ & $0.757\!\pm\!0.019$ \\
\bottomrule
\end{tabular}
}
\end{table}

\section{$k$-NN Representation Robustness}
\label{app:knn-robustness}

\Cref{subsec:cascade-stage1} reports $k$-NN consistency under
standardized Euclidean distance on the full $d{=}38$ state. With
$n_{\text{buffer}} = 800$ and $k{=}10$, that geometry sits near the
regime where neighborhood relationships can become metric-fragile. To
check whether the local-ambiguity diagnosis depends on the metric or
dimensionality, we recompute $k$-NN consistency $\bar c$ and the
held-out $10$-NN selector under three alternative representations of
the same buffer state: the $5$-dimensional base-model probability
subvector alone (\textit{probs-only}), a PCA projection to ten
components fit on the buffer (\textit{pca10}), and Mahalanobis
distance using the buffer covariance with a small ridge
(\textit{mahalanobis}). All projections fit on the buffer only and
are applied to held-out points without leakage.

\begin{center}
\small
\begin{tabular}{lccc}
\toprule
Variant & $\bar c$ & 10-NN oracle agree. & 10-NN test acc. \\
\midrule
Full ($d{=}51$, Euclidean) & $0.387 \pm 0.010$ & $0.491 \pm 0.043$ & $0.749 \pm 0.030$ \\
Probs-only ($d{=}5$)       & $0.431 \pm 0.020$ & $0.477 \pm 0.041$ & $0.762 \pm 0.031$ \\
PCA-10                     & $0.382 \pm 0.010$ & $0.462 \pm 0.030$ & $0.750 \pm 0.025$ \\
Mahalanobis ($d{=}51$)     & $0.430 \pm 0.014$ & $0.503 \pm 0.030$ & $0.738 \pm 0.027$ \\
\bottomrule
\end{tabular}
\end{center}

Across all four representations, $\bar c$ stays in $[0.38, 0.43]$ and
the held-out $10$-NN test accuracy stays in $[0.74, 0.76]$. The
probability-subspace and Mahalanobis variants raise consistency by
roughly five points relative to the full Euclidean baseline, but no
representation breaks above $0.5$ on either consistency or oracle
agreement, and none beats the static reference of $0.762$ on test
accuracy. The Stage-1 diagnosis (that nearby states frequently disagree
on the oracle action) is therefore not a metric artifact. It is robust
to dimensionality reduction, subspace selection, and metric choice.
These numbers are computed on the augmented $d{=}51$ state used in
\Cref{subsec:cascade-stage3}. On the unaugmented $d{=}38$ state, the
full variant matches the body's headline $0.388 \pm 0.010$.

\section{Distribution-Shift Tables}
\label{app:dist-shift}

\Cref{tab:app-dist-shift} reports the full buffer-to-test oracle-shift
summary used for \Cref{fig:oracle-shift}, including the controlled
classifier comparison to the buffer oracle.

\begin{table}[H]
\centering
\caption{Full 16-window buffer-to-test oracle-shift summary. The
``clf$\to$buffer'' column is the controlled classifier's difference in
distance to the buffer oracle versus the test oracle. Positive values mean
the classifier remains closer to the buffer than to deployment.}
\floatlabel{tab:app-dist-shift}
{\footnotesize\setlength{\tabcolsep}{3pt}\renewcommand{\arraystretch}{0.95}%
\begin{tabular}{cc ccc}
\toprule
obs & pred & $d_{\mathrm{TV}}(\pi_\beta, \pi^*)$ & clf agree & clf$\to$buffer \\
\midrule
7 & 7 & $0.060\!\pm\!0.018$ & $0.406\!\pm\!0.044$ & $0.028\!\pm\!0.011$ \\
7 & 14 & $0.079\!\pm\!0.032$ & $0.413\!\pm\!0.043$ & $0.000\!\pm\!0.057$ \\
7 & 21 & $0.064\!\pm\!0.014$ & $0.472\!\pm\!0.039$ & $0.024\!\pm\!0.048$ \\
7 & 28 & $0.083\!\pm\!0.013$ & $0.539\!\pm\!0.032$ & $0.031\!\pm\!0.033$ \\
14 & 7 & $0.048\!\pm\!0.022$ & $0.419\!\pm\!0.048$ & $0.015\!\pm\!0.042$ \\
14 & 14 & $0.070\!\pm\!0.015$ & $0.508\!\pm\!0.041$ & $0.016\!\pm\!0.040$ \\
14 & 21 & $0.049\!\pm\!0.021$ & $0.487\!\pm\!0.043$ & $-0.004\!\pm\!0.038$ \\
14 & 28 & $0.056\!\pm\!0.033$ & $0.512\!\pm\!0.031$ & $0.036\!\pm\!0.042$ \\
21 & 7 & $0.075\!\pm\!0.028$ & $0.430\!\pm\!0.016$ & $0.038\!\pm\!0.021$ \\
21 & 14 & $0.068\!\pm\!0.028$ & $0.464\!\pm\!0.030$ & $0.021\!\pm\!0.035$ \\
21 & 21 & $0.062\!\pm\!0.010$ & $0.482\!\pm\!0.044$ & $0.020\!\pm\!0.030$ \\
21 & 28 & $0.060\!\pm\!0.011$ & $0.412\!\pm\!0.010$ & $0.004\!\pm\!0.039$ \\
28 & 7 & $0.048\!\pm\!0.011$ & $0.451\!\pm\!0.022$ & $0.022\!\pm\!0.019$ \\
28 & 14 & $0.075\!\pm\!0.022$ & $0.465\!\pm\!0.056$ & $0.043\!\pm\!0.030$ \\
28 & 21 & $0.052\!\pm\!0.026$ & $0.448\!\pm\!0.047$ & $0.021\!\pm\!0.032$ \\
28 & 28 & $0.082\!\pm\!0.025$ & $0.469\!\pm\!0.045$ & $0.013\!\pm\!0.048$ \\
\bottomrule
\end{tabular}
}
\end{table}

\section{External-Stratifier Robustness}
\label{app:external-stratifier}

The slices below recompute the main specialization analysis using
stratifiers that do not depend on the base-model probabilities themselves.
The compact summary in \Cref{tab:app-stratifier-summary} distills the full
exported pairwise analysis into the comparisons most relevant to the
workshop submission.

\begin{table}[H]
\centering
\small
\caption{Compact summary of the external-stratifier robustness analysis.
External stratifiers correlate moderately with each other, but their
agreement with the original probability-derived risk-bin construction is
much weaker on the main configuration.}
\floatlabel{tab:app-stratifier-summary}
\begin{tabular}{lcc}
\toprule
\textbf{Stratifier pair} & \textbf{Mean over 16 windows} & \textbf{$(14\text{d},14\text{d})$} \\
\midrule
course\_baserate $\times$ kmeans\_k3 & $0.693$ & $0.687 \pm 0.128$ \\
course\_baserate $\times$ kmeans\_k5 & $0.657$ & $0.654 \pm 0.124$ \\
kmeans\_k3 $\times$ kmeans\_k5 & $0.673$ & $0.641 \pm 0.309$ \\
course\_baserate $\times$ risk\_bin & $0.557$ & $0.315 \pm 0.090$ \\
kmeans\_k3 $\times$ risk\_bin & $0.466$ & $0.343 \pm 0.373$ \\
kmeans\_k5 $\times$ risk\_bin & $0.482$ & $0.298 \pm 0.143$ \\
\bottomrule
\end{tabular}
\end{table}

At $(14\text{d},14\text{d})$, the strongest agreement is between the three
external stratifiers, not between any external stratifier and the original
risk-bin partition. We therefore treat the specialization story as fragile
rather than as a main-paper result.

\subsection{Full 16-window correlation matrix}

\Cref{tab:app-stratifier-corr} reports the per-window correlation matrix
underlying the aggregate robustness summary.

\begin{table}[H]
\centering
\small
\caption{Full 16-window correlation matrix for the external-stratifier
robustness analysis. `cb' denotes course baserate, `rb' denotes the
probability-derived risk bin, and `k3/k5' denote the two clustering-based
stratifiers. The main $(14\text{d},14\text{d})$ setting is unusual in that
all three correlations involving the original risk-bin partition are low.}
\floatlabel{tab:app-stratifier-corr}
\begin{tabular}{cc cccccc}
\toprule
obs & pred & cb$\times$k3 & cb$\times$k5 & cb$\times$rb & k3$\times$k5 & k3$\times$rb & k5$\times$rb \\
\midrule
7 & 7 & 0.607 & 0.656 & 0.708 & 0.606 & 0.451 & 0.492 \\
7 & 14 & 0.605 & 0.538 & 0.519 & 0.666 & 0.344 & 0.534 \\
7 & 21 & 0.823 & 0.767 & 0.524 & 0.884 & 0.517 & 0.473 \\
7 & 28 & 0.543 & 0.437 & 0.458 & 0.739 & 0.586 & 0.571 \\
14 & 7 & 0.758 & 0.702 & 0.711 & 0.602 & 0.467 & 0.666 \\
14 & 14 & 0.687 & 0.654 & 0.315 & 0.641 & 0.343 & 0.298 \\
14 & 21 & 0.691 & 0.643 & 0.217 & 0.746 & 0.381 & 0.115 \\
14 & 28 & 0.756 & 0.688 & 0.283 & 0.693 & 0.100 & 0.187 \\
21 & 7 & 0.655 & 0.689 & 0.840 & 0.648 & 0.499 & 0.670 \\
21 & 14 & 0.711 & 0.675 & 0.579 & 0.642 & 0.660 & 0.551 \\
21 & 21 & 0.708 & 0.796 & 0.475 & 0.801 & 0.283 & 0.442 \\
21 & 28 & 0.778 & 0.743 & 0.499 & 0.611 & 0.466 & 0.309 \\
28 & 7 & 0.604 & 0.710 & 0.835 & 0.628 & 0.291 & 0.635 \\
28 & 14 & 0.708 & 0.730 & 0.755 & 0.688 & 0.708 & 0.708 \\
28 & 21 & 0.810 & 0.545 & 0.787 & 0.453 & 0.721 & 0.409 \\
28 & 28 & 0.640 & 0.535 & 0.411 & 0.715 & 0.638 & 0.652 \\
\bottomrule
\end{tabular}
\end{table}

\subsection{Main-configuration slice winners}

To make that fragility more concrete,
\Cref{tab:app-stratifier-slices} reports which base model is favored by the
oracle within each
external slice of the $(14\text{d},14\text{d})$ configuration, and which
single base model actually attains the best held-out accuracy there.

\begin{table}[H]
\centering
\small
\caption{Main-configuration slice winners for the external-stratifier
analysis. Several slices change which model looks best once the strata are
defined independently of the original probability-derived risk bins, which
is why the specialization claim should be treated cautiously.}
\floatlabel{tab:app-stratifier-slices}
\begin{tabular}{llccc}
\toprule
\textbf{Stratifier} & \textbf{Stratum} & \textbf{Mean $n$} & \textbf{Top oracle model} & \textbf{Best held-out model / acc.} \\
\midrule
risk bin & high & 28.6 & RF & All / 0.698 \\
risk bin & low & 103.2 & GB & Stack / 0.900 \\
risk bin & mid & 68.2 & LR & LR / 0.580 \\
course baserate & high & 55.4 & LR & LR / 0.717 \\
course baserate & low & 39.0 & LR & Stack / 0.901 \\
course baserate & mid & 105.6 & GB & RF / 0.747 \\
kmeans k3 & c0 & 68.2 & GB & CalRF/Stack / 0.762 \\
kmeans k3 & c1 & 69.4 & LR & LR / 0.729 \\
kmeans k3 & c2 & 62.4 & GB & Stack / 0.867 \\
kmeans k5 & c0 & 42.0 & LR & LR / 0.728 \\
kmeans k5 & c1 & 37.8 & GB & Stack / 0.758 \\
kmeans k5 & c2 & 46.4 & GB & CalRF / 0.901 \\
kmeans k5 & c3 & 50.6 & LR & Stack / 0.666 \\
kmeans k5 & c4 & 23.2 & GB & GB / 0.876 \\
\bottomrule
\end{tabular}
\end{table}

\section{Training and Uncertainty Details}
\label{app:training-details}

All main-paper numbers are means and standard deviations over 5 random
seeds. Each seed resamples $1{,}000$ student-course pairs, regenerates
the stratified $800/200$ train-test split, and rebuilds the 4-fold CV
buffer.

The behavioral-cloning model is a two-layer MLP ($d \rightarrow 64
\rightarrow 5$) trained with hard-label cross-entropy for 30 epochs,
batch size 64, dropout 0.2, and weight decay $10^{-4}$. The main-paper
DQN is a one-step contextual-bandit reduction with $\gamma=0$,
architecture ($d \rightarrow 128 \rightarrow 64 \rightarrow 5$), soft
target updates with $\tau=0.05$, and 50 training epochs using the
canonical $\{0,1\}$ oracle-match reward defined in
\Cref{sec:cascade}. The log-probability shaping variant used in
earlier drafts is reported as a reward-sensitivity row in
Appendix~\ref{app:dqn-marginals}. The probabilities-only and
disagreement-augmented variants use the same BC objective with the
corresponding input state.

Paired uncertainty intervals in the body are bootstrap intervals
computed over the 5-seed per-configuration summaries. We use those
intervals only for the comparisons that materially affect the paper's
claims: BC-full versus best static accuracy, probabilities-only BC
versus BC-full accuracy, and disagreement-augmented BC versus BC-full
accuracy.

\end{document}